\documentclass[11pt,a4paper]{article}
\usepackage[hyperref]{acl2021}
\usepackage{times}
\usepackage{latexsym}

\usepackage{multirow}
\usepackage{graphicx}
\usepackage{amsmath}
\usepackage{amsfonts}
\usepackage{amssymb}
\usepackage{amsfonts}
\usepackage{dsfont}
\usepackage{cleveref}
\crefformat{section}{\S#2#1#3} 
\crefformat{subsection}{\S#2#1#3}
\crefformat{subsubsection}{\S#2#1#3}
\usepackage{microtype}

\aclfinalcopy 
\def\aclpaperid{1681} 

\author{Deng Cai$^\heartsuit$, Yan Wang$^\spadesuit$, Huayang Li$^\spadesuit$, Wai Lam$^\heartsuit$, \textnormal{and} Lemao Liu$^\spadesuit$\\
	\\
	$^\heartsuit$The Chinese University of Hong Kong\\
		{\tt thisisjcykcd@gmail.com} \\
		{\tt wlam@se.cuhk.edu.hk} \\
	$^\spadesuit$Tencent AI Lab \\
	{\tt \{brandenwang,alanili,redmondliu\}@tencent.com} \\}

\title{Neural Machine Translation with Monolingual Translation Memory\thanks{~~The work described in this paper is partially supported by a grant from the Research Grant Council of the Hong Kong Special Administrative Region, China (Project Code: 14200719).}}
\date{}
\begin{document}	
	\maketitle
	\begin{abstract}
		Prior work has proved that Translation memory (TM) can boost the performance of Neural Machine Translation (NMT). In contrast to existing work that uses bilingual corpus as TM and employs source-side similarity search for memory retrieval, we propose a new framework that uses monolingual memory and performs learnable memory retrieval in a cross-lingual manner. Our framework has unique advantages. First, the cross-lingual memory retriever allows abundant monolingual data to be TM. Second, the memory retriever and NMT model can be jointly optimized for the ultimate translation goal. Experiments show that the proposed method obtains substantial improvements.  Remarkably, it even outperforms strong TM-augmented NMT baselines using bilingual TM. Owning to the ability to leverage monolingual data, our model also demonstrates effectiveness in low-resource and domain adaptation scenarios.
	\end{abstract}
	\section{Introduction}
	Augmenting parametric neural network models with non-parametric memory \cite{khandelwal2019generalization,guu2020realm,lewis2020pre,lewis2020retrieval} has recently emerged as a promising direction to relieve the demand for ever-larger model size \cite{devlin-etal-2019-bert,radford2019language,Brown2020LanguageMA}. For the task of Machine Translation (MT), inspired by the Computer-Aided Translation (CAT) tools by professional human translators for increasing productivity for decades \cite{yamada2011effect}, the usefulness of Translation Memory (TM) has long been recognized \cite{transmart2021}. In general, TM is a database that stores pairs of source text and its corresponding translations. Like for human translation, early work \cite[][inter alia]{koehn2010convergence,he2010bridging,utiyama2011searching,wang2013integrating} presents translations for similar source input to statistical translation models as additional cues.
	
	Recent work has confirmed that TM can help Neural Machine Translation (NMT) models as well. In a similar spirit to prior work, TM-augmented NMT models do not discard the training corpus after training but keep exploiting it in the test time. These models perform translation in two stages: In the retrieval stage, a retriever searches for nearest neighbors (i.e., source-target pairs) from the training corpus based on source-side similarity such as lexical overlaps \cite{gu2018search,zhang-etal-2018-guiding,xia2019graph}, embedding-based matches \cite{cao-xiong-2018-encoding}, or a hybrid \cite{bulte-tezcan-2019-neural,xu-etal-2020-boosting}; In the generation stage, the retrieved translations are injected into a standard NMT model by attending over them with sophisticated memory networks \cite{gu2018search,cao-xiong-2018-encoding,xia2019graph,he2021fast} or directly concatenating them to the source input \cite{bulte-tezcan-2019-neural,xu-etal-2020-boosting}, or biasing the word distribution during decoding \cite{zhang-etal-2018-guiding}. Most recently, \newcite{khandelwal2020nearest} propose a token-level nearest neighbor search using complete translation context, i.e., both the source-side input and target-side prefix.
	
	Despite their differences, we identify two major limitations in previous research. First, the translation memory has to be a \textit{bilingual} corpus consisting of aligned source-target pairs. This requirement limits the memory bank to bilingual pairs and precludes the use of abundant monolingual data, which can be especially helpful for low-resource scenarios. Second, the memory retriever is \textit{non-learnable}, not end-to-end optimized, and lacks for the ability to adapt to specific downstream NMT models. Concretely, current retrieval mechanisms (e.g., BM25) are \textit{generic} similarity search, adopting a simple heuristic. That is, the more a source sentence overlaps with the input sentence, the more likely its target-side translation pieces will appear in the correct translation. Although this observation is true, the most similar one does not necessarily serve the best for NMT models.
	Ideally, the retrieval metric would be learned from the data in a task-dependent way: we wish to consider a memory only if it can indeed boost the quality of final translation.
	
	In this work, we propose to augment NMT models with \textit{monolingual} TM and a \textit{learnable} \textit{cross-lingual} memory retriever. Specifically, we align source-side sentences and the corresponding target-side translations in a latent vector space using a simple dual-encoder framework \cite{bromley1993signature}, such that the distance in the latent space yields a score function for retrieval. As a result, our memory retriever directly connects the dots between the source-side input and target-side translations, enabling \textit{monolingual} data in the target language to be used alone as TM. Before running each translation, the memory retriever selects the highest-scored memories from a large collection of monolingual sentences (TM), which may include but are not limited to the target side of training corpus, and then the downstream NMT model attends over those memories to help inform its translation. We design the memory retriever with differentiable neural networks. To unify the memory retriever and its downstream NMT model into a learnable whole, the retrieval scores are used to bias the attention scores to the most useful retrieved memories. In this way, our memory retrieval can be end-to-end optimized for the translation objective: a retrieval that improves the golden translation’s likelihood is helpful and should be rewarded, while an uninformative retrieval should be penalized.
	
	One challenge for training our proposed framework is that, when starting from random initialization, the retrieved memories will likely be totally unrelated to the input. Since the memory retriever does not exert positive influence on NMT model's performance, it cannot receive a meaningful gradient and improve. This causes the NMT model to learn to ignore all retrieved memories. To avoid this cold-start problem, we propose to warm-start the retrieval model using two cross-alignment tasks.
	
	Experiments show that (1) Our model leads to significant improvements over non-TM baseline NMT model, even outperforming strong TM-augmented baselines. This is remarkable given that previous TM-augmented models rely on bilingual TM while our model only exploits the target side. (2) Our model can substantially boost translation quality in low-resource scenarios by utilizing extra monolingual TM that is not present in training pairs. (3) Our model gains a strong cross-domain transferability by hot-swapping domain-specific monolingual memory.
	\section{Related Work}
	\begin{figure*}[t]
		\centering
		\includegraphics[width=0.98\textwidth]{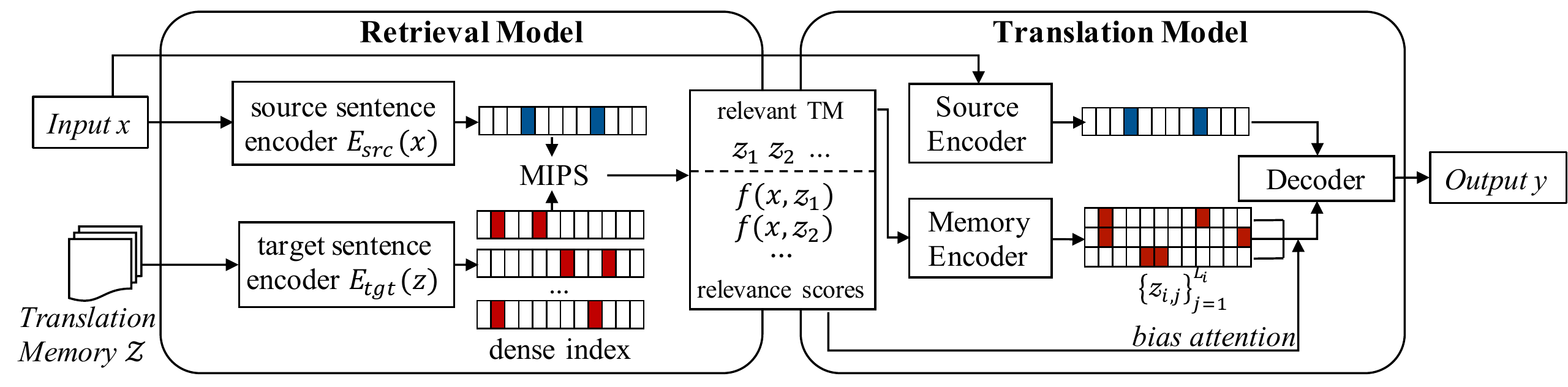}
		\caption{Overall framework. For an input sentence $x$ in the source language, the retrieval model uses Maximum Inner Product Search (MIPS) to find the top-$M$ TM sentences $\{z_i\}^M_{i=1}$  in the target language. The translation model takes $\{z_i\}^M_{i=1}$ and corresponding relevance scores $\{f(x, z_i)\}_{i=1}^M$ as input and generate the translation $y$.}
		\label{arch}
	\end{figure*}
	\paragraph{TM-augmented NMT}
	This work contributes primarily to the research line of Translation Memory (TM) augmented Neural Machine Translation (NMT). \newcite{feng-etal-2017-memory} augmented NMT with a bilingual dictionary to tackle infrequent word translation. \newcite{gu2018search} proposed a model that retrieves examples similar to the test source sentence and encodes retrieved source-target pairs with key-value memory networks. \newcite{cao-xiong-2018-encoding,cao2019learning} used a gating mechanism to balance the impact of the translation memory. \newcite{zhang-etal-2018-guiding} proposed guiding models by retrieving $n$-grams and up-weighting the probabilities of retrieved $n$-grams. \newcite{bulte-tezcan-2019-neural} and \newcite{xu-etal-2020-boosting} used fuzzy-matching with translation memories and augment source sequences with retrieved source-target pairs. \newcite{xia2019graph} directly ignored the source side of a TM and packed the target side into a compact graph. \newcite{khandelwal2020nearest} ran existing translation model on large bi-text corpora and recorded all hidden states for later nearest neighbor search at \textit{each} decoding step, which is very compute-intensive. The distinctions between our work and prior work are obvious: (1) The TM in our framework is a collection of monolingual sentences rather than bilingual sentence pairs; (2) We use learnable task-specific retrieval rather than generic retrieval mechanisms.
	\paragraph{Retrieval for Text Generation}
	Discrete retrieval as an intermediate step has been shown beneficial to a variety of natural language processing tasks. One typical use is to retrieve supporting evidence for open-domain question answering \cite[e.g.,][]{chen-etal-2017-reading,lee-etal-2019-latent,karpukhin-etal-2020-dense}. Recently, retrieval-guided generation has gained increasing interest in a wide range of text generation tasks such as language modeling \cite{guu-etal-2018-generating,khandelwal2019generalization,guu2020realm}, dialogue response generation \cite{weston-etal-2018-retrieve,wu2019response,cai2019skeleton,cai-etal-2019-retrieval}, code generation \cite{hashimoto2018retrieve} and other knowledge-intensive generation \cite{lewis2020retrieval}. It can be observed that there is a shift from using off-the-shelf search engines to learning task-specific retrievers. Our work draws inspiration from this line of research. However, retrieval-guided generation has so far been mainly investigated for knowledge retrieval in the same language. The memory retrieval in this work is more challenging due to the cross-lingual setting.
	\paragraph{NMT using Monolingual Data}
	To our knowledge, the integration of monolingual data for NMT was first investigated by \newcite{gulcehre2015using}, who separately trained target-side language models using monolingual data, and then integrated them during decoding either through re-scoring the beam, or by feeding the hidden state of the language model to the NMT model. \newcite{jean2015montreal} also explored re-ranking the NMT output with a $n$-gram language model. Another successful method for leveraging monolingual data in NMT is \textit{back-translation} \cite{sennrich-etal-2016-improving,fadaee-etal-2017-data,edunov-etal-2018-understanding,he2016dual}, where a reverse translation model is used to translate monolingual sentences from the target language to the source language to generate synthetic parallel sentences. Recent studies \cite{jiao-etal-2021-self,he2019revisiting} showed that \textit{self-training}, where the synthetic parallel sentences are created by translating monolingual sentences in the source language, is also helpful. Our method is orthogonal to previous work and bears a unique feature: it can use more monolingual data without re-training (see \cref{lowres}).
	\section{Proposed Approach}
	\label{approach}
	We start by formalizing the translation task as a retrieve-then-generate process in \cref{method.overivew}. Then in \cref{method.retrieval}, we describe the model design for the cross-lingual memory retrieval model. In \cref{method.translation}, we describe the model design for the memory-augmented translation model. Lastly, we show how to optimize the two components jointly using standard maximum likelihood training  in \cref{method.traninig} and therein we address the cold-start problem via cross-alignment pre-training.
	\subsection{Overview}
	\label{method.overivew}
	Our approach decomposes the whole translation processing into two steps: retrieve, then generate. The overall framework is illustrated in Figure \ref{arch}. The Translation Memory (TM) in our approach is a collection of sentences in the target language $\mathcal{Z}$. Given an input $x$ in the source language, the retrieval model first selects a number of possibly helpful sentences $\{z_i\}_{i=1}^M$ from $\mathcal{Z}$, where $M\ll|\mathcal{Z}|$, according to a relevance function $f(x, z_i)$. Then, the translation model conditions on both the retrieved set $\{(z_i, f(x, z_i)\}_{i=1}^M$ and the original input $x$ to generate the output $y$ using a probabilistic model $p(y|x, z_1, f(x, z_1), \ldots, z_M, f(x, z_M))$. Note that the relevance scores $\{f(x, z_i)\}_{i=1}^M$ are also part of the input to the translation model, encouraging the translation model to focus more on more relevant sentences. During training, maximizing the likelihood of the translation references improves both the translation model and the retrieval model.
	\subsection{Retrieval Model}
	\label{method.retrieval}
	The retrieval model is responsible for selecting the most relevant sentences for a source sentence from a large monolingual TM. This could involve measuring the relevance scores between the source sentence and millions of candidate target sentences, which poses a serious computational challenge. To address this, we implement the retrieval model using a simple dual-encoder framework \cite{bromley1993signature} such that the selection of the most relevant sentences can be reduced to Maximum Inner Product Search (MIPS). With performant data structures and search algorithms \cite[e.g.,][]{shrivastava2014asymmetric,malkov2018efficient}, the retrieval can be done efficiently.
	
	Specifically, we define the relevance score $f(x, z)$ between the source sentence $x$ and the candidate sentence $z$ as the dot product of their dense vector representations:
	\begin{equation}
	f(x, z) = E_{\text{src}}(x)^{\text{T}} E_{\text{tgt}}(z)
	\nonumber
	\end{equation}
	where $E_{\text{src}}$ and $E_{\text{tgt}}$ are the source sentence encoder and the target sentence encoder that map $x$ and $z$ to $d$-dimensional vectors respectively. We implement the two sentence encoders using two independent Transformers \cite{vaswani2017attention}. For an input sentence, we prepend the \texttt{[BOS]} token to its token sequence and then feed it into a Transformer. We take the representation at the \texttt{[BOS]} token as the output (denoted $\text{Trans}_{\{\text{src}, \text{tgt}\}}(\{x, z\})$), and perform a linear projection ($W_{\{\text{src}, \text{tgt}\}}$) to reduce the dimensionality of the vector. Finally, we normalize the vectors to regulate the range of relevance scores.
	\begin{align*}
	E_{\text{src}}(x)  &= \text{normalize}(W_{\text{src}} \text{Trans}_{\texttt{src}}(x)) \\\nonumber
	E_{\text{tgt}}(z) &= \text{normalize}(W_{\text{tgt}}  \text{Trans}_{\texttt{tgt}}(z))	\nonumber
	\end{align*}
	The normalized vectors have zero means and unit lengths. Therefore, the relevance scores always fall in the interval $[-1, 1]$. We let $\theta$ denote all parameters associated with the retrieval model.
	
	In practice, the dense representations of all sentences in TM can be pre-computed and indexed using \textsc{FAISS} \cite{johnson2019billion}, an open-source toolkit for efficient vector search. Given a source sentence $x$ in hand, we compute the vector representation $v_x = E_{\text{src}}(x)$ and retrieve the top $M$ target sentences with vectors closest to $v_x$.
	\subsection{Translation Model}
	\label{method.translation}
	Given a source sentence $x$, a small set of relevant TM $\{z_i\}_{i=1}^M$, and relevance scores $\{f(x, z_i)\}_{i=1}^M$, the translation model defines the conditional probability $p(y|x, z_1, f(x, z_1), \ldots, z_M, f(x, z_M))$.
	
	Our translation model is built upon the standard encoder-decoder NMT model \cite{bahdanau2015neural,vaswani2017attention}: the (source) encoder transforms the source sentence $x$ into dense vector representations. The decoder generates an output sequence $y$ in an auto-regressive fashion. At each time step $t$, the decoder attends over both previously generated sequence $y_{1:t-1}$ and the output of the source encoder, generating a hidden state $h_t$. The hidden state $h_t$ is then converted to next-token probabilities through a linear projection followed by softmax function, i.e., $P_v = \text{softmax}(W_vh_t + b_v)$.
	
	To accommodate the extra memory input, we extend the standard encoder-decoder NMT framework with a memory encoder and allow cross-attention from the decoder to the memory encoder. Specifically, the memory encoder encodes each TM sentence $z_i$ individually, resulting in a set of contextualized token embeddings $\{z_{i, k}\}_{k=1}^{L_i}$, where $L_i$ is the length of the token sequence $z_i$. We compute a cross attention over all TM sentences:
	\begin{gather}
	\alpha_{ij} = \frac{\exp({{h_t}^\text{T} W_m z_{i,j}))}}{\sum_{i=1}^{M} \sum_{k=1}^{L_i} \exp({h_t}^\text{T} W_m z_{i,k})} 
	\label{attn}
	\\
	c_t = W_c \sum_{i=1}^{M} \sum_{j=1}^{L_i} \alpha_{ij} z_{i,j}
	\nonumber
	\end{gather}
	where $\alpha_{ij}$ is the attention score of the $j$-th token in $z_i$, $c_t$ is a weighted combination of memory embeddings, and $W_m$ and $W_c$ are trainable matrices. The cross attention is used twice during decoding. First, the decoder’s hidden state $h_t$ is updated by a weighted sum of memory embeddings, i.e., $h_t = h_t + c_t$. Second, we consider each attention score as a probability of copying the corresponding token \cite{gu-etal-2016-incorporating,see-etal-2017-get}. Formally, the next-token probabilities are computed as:
	\begin{equation}
	p(y_t| \cdot) = (1-\lambda_t)P_v(y_t) + \lambda_t \sum_{i=1}^{M} \sum_{j=1}^{L_i} \alpha_{ij} \mathds{1}_{ z_{ij} =y_t }
	\nonumber
	\end{equation} 
	where $\mathds{1}$ is the indicator function and $\lambda_t$ is a gating variable computed by another feed-forward network $\lambda_t= g(h_t, c_t)$.
	
	Inspired by \newcite{lewis2020pre}, to enable the gradient flow from the translation output to the retrieval model, we bias the attention scores with the relevance scores, rewriting Eq. (\ref{attn}) as:
	\begin{equation}
	\alpha_{ij} = \frac{\exp({{h_t}^\text{T}  W_m z_{i,j} + \beta f(x, z_i))}}{\sum_{i=1}^{M} \sum_{k=1}^{L_i} \exp({h_t}^\text{T} W_m  z_{i,k}+\beta f(x, z_i))}
	\label{bias}
	\end{equation}
	where $\beta$ is a trainable scalar that controls the weight of the relevance scores. We let $ \phi$ denote all parameters associated with the translation model.
	\subsection{Training}
	\label{method.traninig}
	We optimize the model parameters $\theta$ and $\phi$ using stochastic gradient descent on the negative log-likelihood loss function $-\log p(y^*|x, z_1, f(x, z_1), \ldots, z_M, f(x, z_M))$, where $y^*$ refers to the reference translation. As implied by Eq. (\ref{bias}), TM sentences that improve the likelihood of reference translations should receive higher attention scores and higher relevance scores, so gradient descent on the loss function will improve the quality of the retrieval model as well.
	\paragraph{Cross-alignment Pre-training}
	However, if the retrieval model starts from random initialization, all top TM sentences $z_i$ will likely be unrelated to $x$ (or equally useless). This leads to a problem that the retrieval model cannot receive meaningful gradients and improve, and the translation model will learn to completely ignore the TM input. To avoid this cold-start problem, we propose two cross-alignment tasks to warm-start the retrieval model. 
	
	The first task is sentence-level cross-alignment. This task aims to find the right translation for a source sentence given a set of other translations, which is directly related to our retrieval function. Concretely, We sample $B$ source-target pairs from the training corpus at each training step. Let $X$ and $Z$ be the $(B\times d)$ matrix of the source and target vectors encoded by $E_{\text{src}}$ and $E_{\text{tgt}}$ respectively. $S=XZ^{T}$ is a $(B\times B)$ matrix of relevance scores, where each row corresponds to a source sentence and each column corresponds to a target sentence. Any $(X_i, Z_j)$ pair should be aligned when $i=j$, and should not otherwise. The objective is to maximize the scores along the diagonal of the matrix and henceforth reduce the values in other entries. The loss function can be written as:
	\begin{equation}
	\mathcal{L}_{\texttt{snt}}^{(i)} =\frac{-\exp(S_{ii})}{\exp(S_{ii}) + \sum_{j\ne i}\exp(S_{ij})}. \nonumber
	\end{equation}
	
	The second task is token-level cross-alignment, which aims to predict the tokens in the target language given the source sentence representation and vice versa. Formally, we use bag-of-words losses:
	\begin{eqnarray}
	\mathcal{L}_{\texttt{tok}}^{(i)} =-\sum_{w_y\in \mathcal{Y}_i} \log p(w_y | X_i ) + \sum_{w_x\in \mathcal{X}_i} \log p(w_x | Y_i ) \nonumber
	\end{eqnarray}
	where $\mathcal{X}_i$ ($\mathcal{Y}_i$) represents the set of tokens in the $i$-th source (target) sentence and the token probabilities are computed by a linear projection followed by the softmax function. The joint loss for pre-training is $\frac{1}{B}\sum_{i=1}^B\mathcal{L}_{\texttt{snt}}^{(i)}  + \mathcal{L}_{\texttt{tok}}^{(i)}$. In practice, we find that both the sentence-level and token-level objectives are crucial for achieving superior performance. 
	\paragraph{Asynchronous Index Refresh}
	To employ fast MIPS, we must pre-compute $E_{\text{tgt}}(z)$ for every $z\in \mathcal{Z}$ and build an index. However, the index cannot remain consistent with the running model during training as $\theta$ will be updated over time. One straightforward solution to fix the parameters of $E_{\text{tgt}}$ after the pre-training described above and only fine-tune the parameters of $E_{\text{src}}$. However, this may hurt performance since $E_{\text{tgt}}$ cannot adapt to the translation objective. Another solution is to asynchronously refresh the index by re-computing and re-indexing all TM sentences at regular intervals. The index is slightly outdated between refreshes, however, we use fresh $E_{\text{tgt}}$ in gradient estimate. We explore both options in our experiments.
		\begin{table}[t]
		\centering
		\small
		\begin{tabular}{l|c|c|c}
			\hline
			Dataset & \#Train Pairs& \#Dev Pairs& \#Test Pairs\\
			\hline
			{\bf En$\Leftrightarrow$Es}  &679,088&2,533&2,596\\
			{\bf En$\Leftrightarrow$De} &699,569&2,454&2,483 \\
			\hline
		\end{tabular}
		\caption{Data statistics for the JRC-Acquis corpus.}
		\label{statistics}
	\end{table}
	\section{Experiments}
		\begin{table*}[t]
		\centering
		\small
		\begin{tabular}{l|l|l||ll|ll|ll|ll}
			\hline
			\multirow{2}{*}{\bf \#}   & \multirow{2}{*}{\bf System}  &   \multirow{2}{*}{\bf Retriever}  & \multicolumn{2}{c}{\bf Es$\Rightarrow$En}  &  \multicolumn{2}{|c}{\bf En$\Rightarrow$Es} & \multicolumn{2}{|c}{\bf De$\Rightarrow$En} & \multicolumn{2}{|c}{\bf En$\Rightarrow$De} \\
			\cline{4-11}
			&&   &   Dev    &   Test   &  Dev    &   Test  &  Dev    &   Test &  Dev    &   Test \\
			\hline \hline
			\multicolumn{11}{c}{{\em Existing NMT systems*   }} \\
			\hline
			&\newcite{gu2018search} &  source similarity                       &  63.16  & 62.94  &  -  & -  &  -  &  - & -  & -\\
			&\newcite{zhang-etal-2018-guiding}&  source similarity        &  63.97  & 64.30 & 61.50 &  61.56 &  60.10  &  60.26 &55.54  & 55.14\\
			&\newcite{xia2019graph}  & source similarity                        &  66.37  & 66.21  & 62.50  & 62.76 &  61.85  & 61.72 & 57.43  & 56.88\\
			\hline\hline
			\multicolumn{11}{c}{{\em Our NMT systems}}   \\ \hline
			1&\multirow{5}{*}{\em ~~~~this work}  &  None 
			& 64.25 & 64.07 &62.27  &  61.54  & 59.82  & 60.76 & 55.01  & 54.90 \\
			2&& source similarity                          & 66.98 & 66.48 & 63.04  & 62.76  & 63.62  & 63.85  & 57.88  & 57.53 \\
			3&& cross-lingual (fixed)	                   & 66.68 & 66.24 &63.06 & 62.73  & 63.25  & 63.06 & 57.61  & 56.97 \\
			4&&  cross-lingual (fixed $E_{\text{tgt}}$)$\dagger$       & 67.66 & 67.16 &63.73   & 63.22  & 64.39  & 64.01 & 58.12  & 57.92 \\
			5&&  cross-lingual$\dagger$      & \textbf{67.73} &\textbf{67.42}  & \textbf{64.18}  & \textbf{63.86}  & \textbf{64.48}  & \textbf{64.62} & \textbf{58.77} & \textbf{58.42} \\
			\hline
		\end{tabular}
		\caption{Experimental results (BLEU scores) on four translation tasks. $^*$Results are from \newcite{xia2019graph}. $\dagger$The two variants of our method (model \#4 and model \#5) are significantly better than other baselines with $p$-value $<$ 0.01, tested by bootstrap re-sampling \cite{koehn2004statistical}.}
		\label{res-conventional}
	\end{table*}
	We experiment with the proposed approach in three settings: (1) the conventional setting where the available TM is limited to the bilingual training corpus, (2) the low-resource setting where bilingual training pairs are scarce but extra monolingual data is exploited as additional TM, and (3) non-parametric domain adaptation using monolingual TM. Note that existing TM-augmented NMT models are only applicable to the first setting, the last two settings only become possible with our proposed model. We use BLEU score \cite{papineni2002bleu} as the evaluation metric.
	\subsection{Implementation Details}
	We build our model using Transformer blocks with the same configuration as Transformer Base \cite{vaswani2017attention} (8 attention heads, 512 dimensional hidden state, and 2048 dimensional feed-forward state). The number of Transformer blocks is 3 for the retrieval model, 4 for the memory encoder in the translation model, and 6 for the encoder-decoder architecture in the translation model. We retrieve the top 5 TM sentences. The \textsc{FAISS} index code is ``IVF1024\_HNSW32,SQ8" and the search depth is 64.
	
	We follow the learning rate schedule, dropout and label smoothing settings described in \newcite{vaswani2017attention}. We use Adam optimizer \cite{kingma2014adam} and train models with up to 100K steps throughout all experiments. When trained with asynchronous index refresh, the re-indexing interval is 3K training steps.\footnote{Our code is released at \url{https://github.com/jcyk/copyisallyouneed}.}
	\subsection{Conventional Experiments}
	\label{conventional}
	Following prior work in TM-augmented NMT, we first conduct experiments in a setting where the bilingual training corpus is the only source for TM.
	\paragraph{Data}
	We use the JRC-Acquis corpus \cite{steinberger2006jrc} for our experiments. The JRC-Acquis corpus contains the total body of European Union (EU) law applicable to the EU member states. This corpus was also used by \newcite{gu2018search,zhang-etal-2018-guiding,xia2019graph} and we managed to get the datasets originally preprocessed by \newcite{gu2018search}, making it possible to fairly compare our results with previously reported BLEU scores. Specifically, we select four translation directions, namely, Spanish$\Rightarrow$English (Es$\Rightarrow$En), En$\Rightarrow$Es, German$\Rightarrow$English (De$\Rightarrow$En), and En$\Rightarrow$De, for evaluation. Detailed data statistics are shown in Table \ref{statistics}.
	\paragraph{Models}
	To study the effect of each model component, we implement a series of model variants (model \#1 to \#5 in Table \ref{res-conventional}).
	\begin{enumerate}
		\item NMT without TM. To measure the help from TM, we remove the model components related to TM (including the retrieval model and the memory encoder), and only employ the encoder-decoder architecture for NMT. The resulted model is equivalent to the Transformer Base model \cite{vaswani2017attention}.
		\item TM-augmented NMT using source similarity search. To isolate the effect of architectural changes in NMT models, we replace our cross-lingual memory retriever with traditional source-side similarity search. Specifically,  we use the fuzzy match system used in \newcite{xia2019graph} and many others, which is based on BM25 and edit distance.
		\item TM-augmented NMT using pre-trained cross-lingual retriever. To study the effect of end-to-end task-specific optimization of the retrieval model, we pre-train the retrieval model using the cross-alignment tasks introduced in \cref{method.traninig} and keep it fixed in the following NMT training.
		\item Our full model using a fixed TM index; After pre-training, we fix the parameter of $E_{\text{tgt}}$ during NMT training. 
		\item Our full model trained with asynchronous index refresh.
	\end{enumerate}
	\paragraph{Results}
	\begin{figure*}[t]
		\centering
		\includegraphics[width=0.8\textwidth]{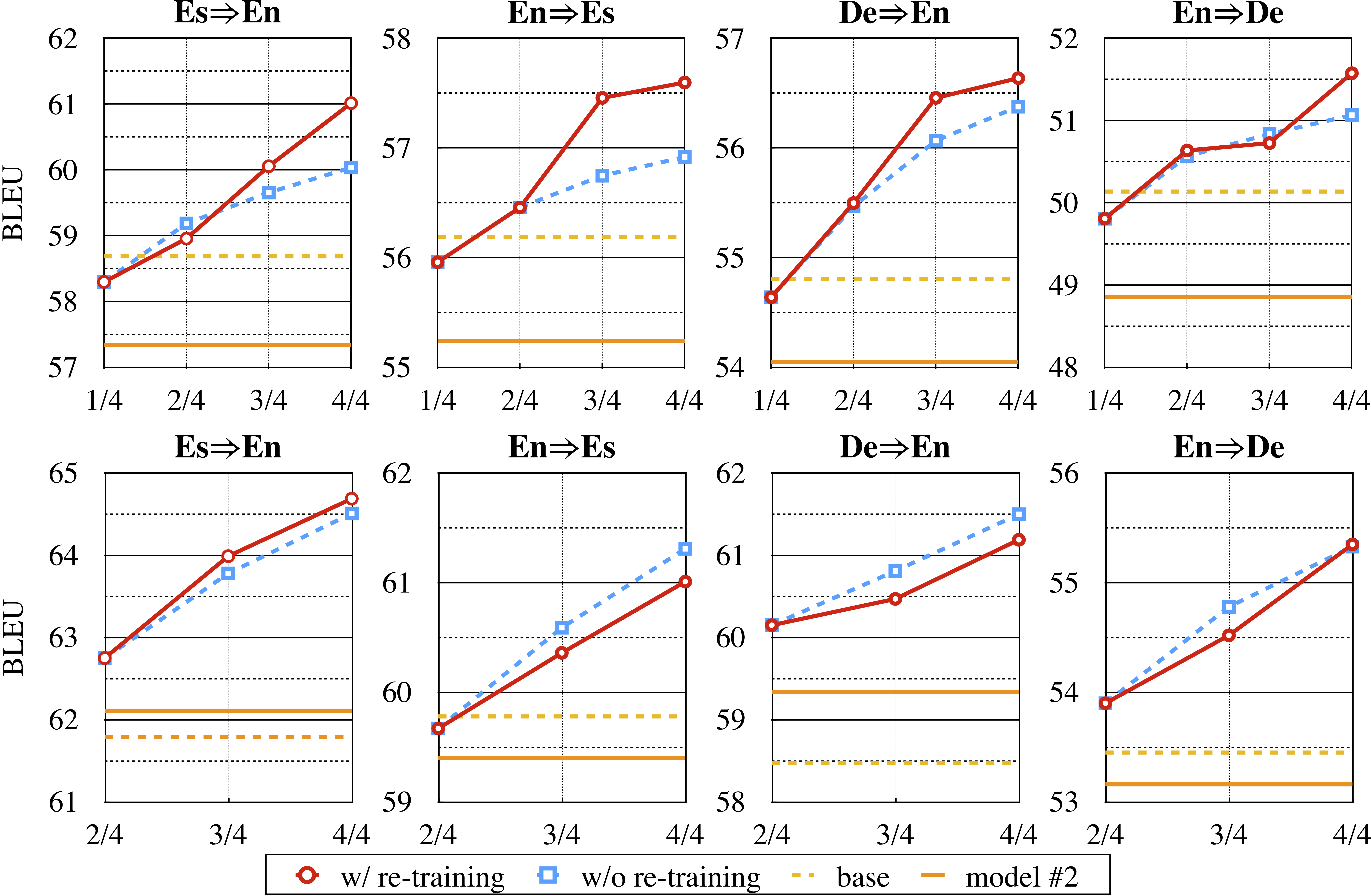}
		\caption{Test results with 1/4 bilingual pairs (upper) and 2/4 bilingual pairs (lower) across different TM sizes.}
		\label{res-lowres}
	\end{figure*}
	The results of the above models are presented in Table \ref{res-conventional}. We have the following observations: (1) Our full model trained with asynchronous index refresh (model \#5) delivers the best performance on test sets across all four translation tasks, outperforming the non-TM baseline (model \#1) by 3.26 BLEU points in average and up to 3.86 BLEU points (De$\Rightarrow$En). This result confirms that monolingual TM can boost NMT performance; (2) The end-to-end learning of the retriever model is the key for substantial performance improvement. We can see that using a pre-trained fixed cross-lingual retriever only gives moderate test performance, fine-tuning $E_{\text{src}}$ and fixing $E_{\text{tgt}}$ significantly boosts the performance, and  fine-tuning both $E_{\text{src}}$ and $E_{\text{tgt}}$ leads to the strongest performance (model \#5$>$model \#4$>$model \#3); (3) Cross-lingual retrieval (model \#4 and model \#5) can obtain better results than that of the source similarity search (model \#2). This is remarkable since the cross-lingual retrieval only requires monolingual TM, while the source similarity search relies on bilingual TM. We attribute the success, again, to the end-to-end adaptability of our cross-lingual retriever. This is manifested by the fact that model \#3 even slightly underperforms model \#2 in some of translation tasks.
	\paragraph{Contrast to Previous Bilingual TM Systems}
	We also compare our results with the best previously reported models.\footnote{Some recent work used different datasets other than JRC-Acquis with unspecified data split, which makes it hard to make an exhaustive comparison. However, note that our in-house baseline (model \#2) is quite strong.} We can see that our results significantly outperform previous arts. Notably, our best model (model \#5) surpasses the best reported model \cite{xia2019graph} by 1.69 BLEU points in average and up to 2.9 BLEU points (De$\Rightarrow$En). This result verifies the effectiveness of our proposed models. In fact, we can see that our translation model using traditional similarity search (model \#2) already outperforms the best previously reported results, which reveals that the architectural design of our translation model is surprisingly effective despite its simplicity.
	\begin{table*}[t]
		\centering
		\small
		\begin{tabular}{l|c|c|c|c|c|c|c|c|c}
			\hline
			\multirow{2}{*}{\bf Data} &\multirow{2}{*}{\bf Model}&\multicolumn{2}{|c}{{\bf Es$\Rightarrow$En}} &\multicolumn{2}{|c}{{\bf En$\Rightarrow$Es}}&  \multicolumn{2}{|c}{{\bf De$\Rightarrow$En}} & \multicolumn{2}{|c}{\bf En$\Rightarrow$De} \\
			\cline{3-10}
			&&dev &test &dev &test &dev &test &dev &test\\
			\hline
			\hline
			\multirow{3}{*}{\shortstack[l]{1/4 bilingual + \\4/4 monolingual}}&Ours&61.46&61.02&57.86&57.40&56.77&56.54&51.11& 51.58\\
			&BT&62.47&61.99&60.28&59.59&57.75&58.20&52.47&52.96\\
			&Ours+BT&\textbf{65.98}&\textbf{65.51}&\textbf{62.48}&\textbf{62.22}&\textbf{62.22}&\textbf{61.79}&\textbf{56.75}&\textbf{56.50}\\
			\hline\hline
			\multirow{3}{*}{\shortstack[l]{2/4 bilingual + \\4/4 monolingual}}&Ours&65.17&64.69&61.31&61.01&61.43&61.19&55.55&55.35\\
			&BT&63.82&63.10&61.59&60.83&59.17&59.26&54.18&54.29\\
			&Ours+BT&\textbf{66.95}&\textbf{66.38}&\textbf{63.22}&\textbf{62.90}&\textbf{63.68}&\textbf{63.10}&\textbf{57.69}&\textbf{57.40}\\
			\hline
		\end{tabular}
		\caption{Comparison with back-translation (BT).}
		\label{res-bt}
	\end{table*}
	\begin{table*}[t]
		\small
		\centering
		\begin{tabular}{l|c|c|c|c|c|c|c}
			\hline
			& {\bf Medical} & {\bf Law} & {\bf IT} & {\bf Koran} & {\bf Subtitle} & {\bf Avg.} & {\bf Avg. $\Delta$}\\
			\hline
			\#Bilingual Pairs&61,388&114,930&55,060&4,458&124,992&-&-\\
			\#Monolingual Sents&184,165&344,791&165,181&13,375&374,977&-&-\\
			\hline\hline
			\multicolumn{8}{c}{Using Bilingual Pairs Only} \\
			\hline
			Transformer Base&47.81&51.40&33.90&14.64&21.64&33.88&-\\
			Ours &47.52&51.17&34.64&15.49&22.66&34.30& +0.42\\
			\hline\hline
			\multicolumn{8}{c}{+ Monolingual Memory} \\
			\hline\hline
			Ours + domain-specific  &\textbf{50.32}&53.97&\textbf{35.33}&\textbf{16.26}&\textbf{22.78}&\textbf{35.73}&\textbf{+1.85} \\
			Ours + all-domains &50.23&\textbf{54.12}&35.24&16.24&\textbf{22.78}&35.72& +1.84\\
			
			\hline
		\end{tabular}
		\caption{Test results on domain adaptation.}
		\label{res-domain}
	\end{table*}
	\subsection{Low-Resource Scenarios}
	\label{lowres}
	One most unique characteristic of our proposed model is that it uses monolingual TM. This motivates us to conduct experiments in low-resource scenarios, where we use extra monolingual data in the target language to boost translation quality.
	\paragraph{Data}
	We create low-resource scenarios by randomly partitioning each training set in JRC-Acquis corpus into four subsets of equal size. We set up two series of experiments: (1) We only use the bilinguals pairs in the first subset and gradually enlarge the TM by including more monolingual data in other subsets. (2) Similar to (1), but we instead use the bilingual pairs in the first two subsets.
	\paragraph{Models}
	As shown in \cref{conventional}, the model trained with asynchronous index refresh (model \#5) is slightly better than the model using fixed $E_{\text{tgt}}$ (model \#4), however, the computational cost of training model \#5 is much bigger. For simplicity and environmental consideration, we only test model \#4 in low-resource scenarios. Nevertheless, we note there are still two modeling choices: (1) train the model once with the TM limited to training pairs and only enlarge the TM during testing; (2) re-train the model with every enlarged TM. Note that when using the first choice, the model may retrieve a TM sentence that has never been seen during training. To measure the performance improvements from additional monolingual TM, we also include a Transformer Base baseline (model \#1, denoted as base) and a bilingual TM baseline (model \#2).
	\paragraph{Results}
	Figure \ref{res-lowres} shows the main results on the test sets. The general patterns are consistent across all experiments: the larger the TM becomes, the better translation performance the model achieves. When using all available monolingual data (4/4), the translation quality is boosted significantly. Interestingly, the performance of models without re-training is comparable to, if not better than, those with re-training. We also observe that when the training pairs are very scarce (only 1/4 bilingual pairs are available), a small size of TM even hurts the model performance. The reason could be overfitting. We speculate that better results would be obtained by tuning the model hyper-parameters according to different TM sizes.
	\paragraph{Contrast to Back-Translation}
	We compare our models with back-translation (BT) \cite{sennrich-etal-2016-improving}, a popular way of utilizing monolingual data for NMT. We train a target-to-source Transformer Base model using bilingual pairs and use the resultant model to translate monolingual sentences to obtain additional synthetic parallel data. As shown in Table \ref{res-bt}, our method performs better than BT with 2/4 bilingual pairs but performs worse with 1/4 bilingual pairs. Interestingly, the combination of BT and our method yields significant further gains, which demonstrates that our method is not only orthogonal but also complementary to BT. 
	\subsection{Non-parametric Domain Adaptation}
	Lastly, the \textit{``plug and play"} property of TM further motivates us to domain adaptation, where we adapt a \textit{single} general-domain model to a specific domain by using domain-specific monolingual TM.
	\paragraph{Data}
	To simulate a diverse multi-domain setting, we use the data splits in \newcite{aharoni-goldberg-2020-unsupervised} originally collected by \newcite{koehn2017six}. It includes German-English parallel data for train/dev/test sets in five domains: Medical, Law, IT, Koran and Subtitles. Similar to the experiments in \cref{lowres}, we only use one fourth of bilingual pairs for training. The target side of the remaining data is treated as additional monolingual data for building domain-specific TM, and the source side is discarded. The data statistics can be found in the upper block of Table \ref{res-domain}. The dev and test sets for each domain contains 2K instances.
	\paragraph{Models}
	We first train a Transformer Base baseline (model \#1) on the concatenation of bilingual pairs in all domains. As in \cref{lowres}, we train our model using fixed $E_{\text{tgt}}$ (model \#4). One advantage of our approach is the possibility of training a single model which can be adapted to any new domain at the inference time without any re-training, by just switching the TM.  When adapting to a new TM, we do not re-train our model. As the purpose here is to verify that our approach can tackle domain adaptation \textit{without any domain-specific training}, we leave the comparison and combination of other domain adaptation techniques \cite{moore-lewis-2010-intelligent,chu-wang-2018-survey} as future work.
	\paragraph{Results}
	The results are presented in Table \ref{res-domain}. We can see that when only using the bilingual data, the TM-augmented model obtains higher BLEU scores in domains with less data but slightly lower scores in other domains compared to the non-TM baseline. However, as we switch the TM to domain-specific TM, the translation quality is significantly boosted in all domains, improving the non-TM baseline by an average of 1.85 BLEU points, with improvements as large as 2.57 BLEU points on Law and 2.51 BLEU point on Medical. We also attempt to combine all domain-specific TMs to one and use it for all domains (the last row in Table \ref{res-domain}). However, we do not obtain noticeable improvement. This reveals that the out-of-domain data can provide little help so that a smaller in-domain TM is sufficient, which is also confirmed by the fact that about 90.21\% of the retrieved sentences come from the corresponding domain in the combined TM.
	\subsection{Running Speed}
	With the help of FAISS in-GPU index, search over millions of vectors can be made incredibly efficient (often in tens of milliseconds). In our implementation, the memory search performs even faster than naive BM25\footnote{Elasticsearch Implementation: \url{https://www.elastic.co/}}. For the results in Table \ref{res-conventional}, taking the vanilla Transformer Base model (model \#1) as the baseline. The inference latency of our models (both model \#4 and model \#5) is about 1.36 times of the baseline (all use a single Nividia V100 GPU). Note that the corresponding number for the previous state-of-the-art model \cite{xia2019graph} is 1.80. As for training cost, the averaged time cost per training step of model \#4 and model \#5 is 2.62 times and 2.76 times of the baseline respectively, which are on par with traditional TM-augmented baselines (model \#2 is 2.59 times) (all use two Nividia V100 GPUs). Table \ref{speed} presents the results. In addition, we also observe that memory-augmented models converge much faster than vanilla models in terms of training steps.
	\section{Conclusion}
	We introduced an effective approach that augments NMT models with monolingual TM. We show that a task-specific cross-lingual memory retriever can be learned by end-to-end MT training. Our approach achieves new state-of-the-art results on several datasets, leads to large gains in low-resource scenarios where the bilingual data is limited, and can specialize a NMT model for specific domains without further training.
	
	Future work should aim to build over our proposed framework. Two obvious directions are: (1) Even though our experiments validated that the whole framework can be learned from scratch using standard MT corpora, it is possible to initialize each model component in our framework with massively pre-trained models for performance enhancement; and (2) The NMT model can benefit from aggregating over a set of diverse memories, which is not explicitly encouraged in current design.
			\begin{table}[t]
		\centering
		\small
		\begin{tabular}{l|c|c|c}
			\hline
			\# & Model & Training & Inference\\
			\hline
			1  &Transformer Base&1.00x&1.00x\\
			2 &source similarity&2.59x& - \\
			4 & cross-lingual (fixed $E_{\text{tgt}}$)&2.62x&1.36x \\
			5 &cross-lingual&2.76x&1.36x \\
			- &\newcite{xia2019graph} & - & 1.80x \\
			\hline
		\end{tabular}
		\caption{Latency cost for training and inference. For training, we measure the averaged time cost per training step. The number of \newcite{xia2019graph} is inferred from their paper.}
		\label{speed}
	\end{table}
	\bibliography{acl2021}
	\bibliographystyle{acl_natbib}
\end{document}